%% file: article.tex
\documentclass[lettersize,journal]{IEEEtran}
\usepackage{amsmath,amsfonts}
\usepackage{algorithmic}
\usepackage{algorithm}
\usepackage{array}
\usepackage[caption=false,font=normalsize,labelfont=sf,textfont=sf]{subfig}
\usepackage{textcomp}
\usepackage{stfloats}
\usepackage{url}
\usepackage{verbatim}
\usepackage{graphicx}
\usepackage{cite}
\usepackage{threeparttable}
\usepackage{multirow}
\usepackage[colorlinks,linkcolor=blue]{hyperref}
\usepackage{orcidlink}
\usepackage{booktabs} 
\usepackage{pifont}
\begin{document}

\title{Plane2Depth: Hierarchical Adaptive Plane Guidance \\ for Monocular Depth Estimation}

\author{
    Li Liu\textsuperscript{$^*$}\orcidlink{0009-0004-3280-8490},
    Ruijie Zhu\textsuperscript{$^*$}\orcidlink{0000-0001-6092-0712}, 
    Jiacheng Deng~\orcidlink{0000-0003-2838-0378},
    Ziyang Song~\orcidlink{0009-0009-6348-8713},
    Wenfei Yang~\orcidlink{0000-0003-3599-7659},
    Tianzhu Zhang\textsuperscript{$^\dagger$ }\orcidlink{0000-0003-0764-6106}
\thanks{$^*$ Equal contribution. $^\dagger$ Corresponding author.}
\thanks{The authors are with Deep Space Exploration Laboratory/School of Information Science and Technology, University of Science and Technology of China (USTC), Hefei 230026, P.R.China. 
Contact: 
\{liu\_li, ruijiezhu, dengjc, songziyang, yangwf\}@mail.ustc.edu.cn;
tzzhang@ustc.edu.cn.}
\thanks{This work has been submitted to the IEEE for possible publication.
Copyright may be transferred without notice, after which this version may
no longer be accessible.}
}




\maketitle

\begin{abstract}
Monocular depth estimation aims to infer a dense depth map from a single image, which is a fundamental and prevalent task in computer vision. 
Many previous works have shown impressive depth estimation results through carefully designed network structures, but they usually ignore the planar information and therefore perform poorly in low-texture areas of indoor scenes.
In this paper, we propose Plane2Depth, which adaptively utilizes plane information to improve depth prediction within a hierarchical framework. 
Specifically, in the proposed plane guided depth generator (PGDG), we design a set of plane queries as prototypes to softly model planes in the scene and predict per-pixel plane coefficients. 
Then the predicted plane coefficients can be converted into metric depth values with the pinhole camera model. 
In the proposed adaptive plane query aggregation (APGA) module, we introduce a novel feature interaction approach to improve the aggregation of multi-scale plane features in a top-down manner. 
Extensive experiments show that our method can achieve outstanding performance, especially in low-texture or repetitive areas. 
Furthermore, under the same backbone network, our method outperforms the state-of-the-art methods on the NYU-Depth-v2 dataset, achieves competitive results with state-of-the-art methods KITTI dataset and can be generalized to unseen scenes effectively.
\end{abstract}


\section{Introduction}
\IEEEPARstart{M}{onocular} depth estimation aims to predict dense depth map from a single image, playing a crucial role in achieving real-world 3D perception and understanding. It is indispensable in various applications such as autonomous driving~\cite{labayrade2002real}, augmented reality~\cite{yucel2021real}, and robot navigation~\cite{metni2005visual}.
Compared to other methods such as MVS~\cite{schonberger2016pixelwise}, monocular depth estimation is cost effective and easy to deploy, requiring only a single RGB image as input.
However, owing to the scale ambiguity of the monocular depth estimation task, where objects appear smaller in the distance and larger up close, a single 2D image can correspond to infinite 3D scenes. This ill-posed nature presents a formidable challenge for monocular depth estimation.

\begin{figure}[ht]
  \centering 
  \includegraphics[width=\linewidth]{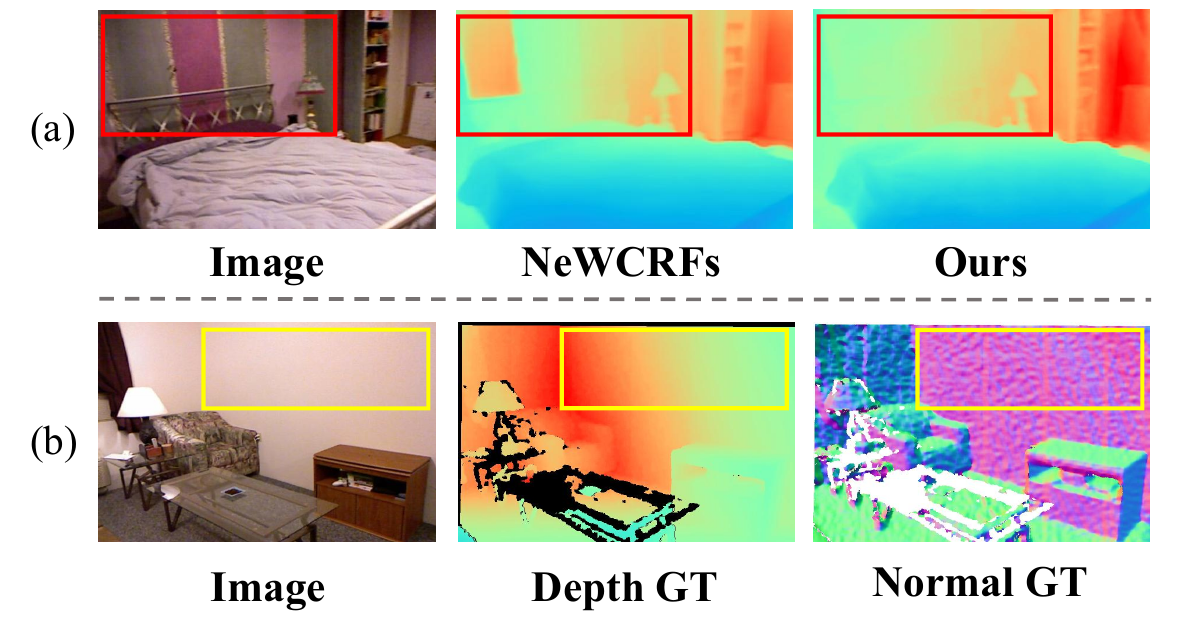} 
  \caption{\textbf{Illustration of our motivation.} We present an example of ``visual deception" in (a). The significant color discrepancies mislead the network into predicting an incorrect depth map. Our method successfully mitigates this issue by using plane information. 
  In (b), pixels within the yellow bounding box correspond to different depths but share the same surface normal (indicated by identical colors representing the corresponding values). Since the generation of ground truth for surface normal depends on conventional algorithms, there exists a slight offset.} 
  \label{fig:1} 
  \vspace{-0.5em}
\end{figure}

To address this challenge, a series of learning-based methods have been proposed that leverage massive depth annotations to drive models to alleviate the scale ambiguity of monocular cameras.
Generally, these methods can be roughly categorized into two main types: 
(1) Continuous regression. These methods~\cite{lee2019big,DPT} directly regress per-pixel corresponding depth. However, the continuous-depth output space may slow down network convergence. 
(2) Discrete regression. The recent methods~\cite{bhat2021adabins,li2022binsformer} essentially transform the regression problem into a classification problem, which alleviates the difficulty of direct depth regression and benefits the network convergence. 
For example, AdaBins~\cite{bhat2021adabins} adaptively divides the depth range of the scene into discrete bins and then predicts the probability that pixels belong to each bin. 
Subsequent methods~\cite{li2022binsformer,shao2023iebins} design different bin division strategies to improve the adaptability to different scene depth distributions.
Although existing methods have demonstrated outstanding performance in multiple benchmarks, they often perform poorly in areas with weak texture and repetitive patterns. 
Therefore, some methods attempt to introduce additional geometric constraints into monocular depth estimation, especially by leveraging plane priors.
For example, BTS~\cite{lee2019big} assumes the neighborhood of pixels forms a local plane and applies local plane assumptions from coarse to fine. However, the local-plane assumption is not always true and may introduce incorrect geometric priors. P3Depth~\cite{patil2022p3depth} constructs coplanar relationships between pixels, but the network may struggle to learn plane priors without explicit planar constraints.
Recently, NDDepth~\cite{shao2023nddepth} uses extra prediction heads to directly predict planes corresponding to pixels, but the multi-branch design makes the network require a complex optimization process.

Based on the discussion above, we summarize two key points that need to be carefully considered to help the network learn planar priors from complex and diverse scenes:
(1) \textbf{How to effectively utilize the planes in the scene to guide depth estimation.}
As shown in Figure~\ref{fig:1} (a), the discontinuity of RGB colors on the wall surface misleads the network to predict incorrect depth. 
This kind of ``visual deception" is difficult to distinguish with only depth supervision. 
\textcolor{black}{The existing methods are limited by the naive planar assumption or the lack of effective planar constraints.}
Furthermore, real 3D scenes are usually complicated and contain planes of varying sizes. 
Therefore, effectively modeling the planes in the scene and constructing corresponding constraints is crucial for depth estimation networks.
(2) \textbf{How to adaptively aggregate planar features of the scene.}
In comparison to depth, planes often exhibit stronger correlations in real-world scenes. 
As shown in Figure~\ref{fig:1} (b), pixels corresponding to the same planar region may display different depths while sharing identical surface normal. 
Based on this observation, we argue that aggregating the relevant features within plane regions can benefit the plane prediction. 
Previous local plane assumptions can be viewed as a simplistic aggregation pattern. 
However, plane regions vary in size and shape in real scenes, presenting challenges in covering all potential cases.
Therefore, it is necessary to design an adaptive feature aggregation approach to assist plane prediction.

To address the aforementioned issues, we propose Plane2Depth, a plane guided hierarchical adaptive framework for monocular depth estimation, which includes a plane guided depth generator with an adaptive plane query aggregation module.
\textbf{In the plane guided depth generator}, we first design a set of plane queries as plane prototypes. 
Then the plane queries interact with multi-scale image features to predict the plane coefficients of each pixel, which can be converted into depth through camera intrinsic.
Specifically, we predict the coefficients of a predetermined number of plane bases through plane queries, and then estimate the probability of pixels belonging to each plane basis. 
By weighting the plane coefficients according to the probabilities, we can predict the coefficients for each pixel, and then calculate the depth map through the camera intrinsic.
\textbf{In the adaptive plane query aggregation module}, we introduce a novel feature interaction approach for top-down query aggregation. 
For each scale, we incorporate an additional transformer layer to modulate the current image features with the help of plane queries from the previous layer. 
This design enables the current image features to adapt to the plane bases of the previous layer, emphasizing the guidance of global contextual features and ensuring the consistency of multi-scale features in query aggregation for better plane prediction.
In summary, our main contributions are as follows:

\begin{itemize}
    \item We propose a novel plane guided hierarchical adaptive framework for monocular depth estimation called Plane2Depth, which consists of a plane guided depth generator and an adaptive plane query aggregation module.
    \item The proposed plane guided depth generator equips the model with planar perception capabilities, enhancing the performance in regions with weak textures and other challenging areas, and also maintaining satisfactory performance in non-planar regions. Furthermore, the proposed adaptive planar query aggregation module ensures consistent multi-scale planar feature aggregation through feature modulation.
    \item Experiments demonstrate that our method outperforms the state-of-the-art methods on the NYU-Depth-v2 dataset, achieves competitive results with state-of-the-art methods on KITTI dataset under the same settings, and exhibits satisfactory generalization ability in the zero-shot test on the SUN RGB-D dataset.
\end{itemize}

\section{Related Works}\label{sec2}
In this section, we succinctly introduce previous research on supervised monocular depth estimation and geometric constraints for depth estimation.
\subsection{Supervised Monocular Depth Estimation}
\textcolor{black}{
Based on the network architecture, we broadly classify supervised monocular depth estimation methods into two categories: continuous regression based and discrete regression based methods.}
Continuous Regression based methods~\cite{bodis2014fast,eigen2014depth,sufaceNormal,forest,saxena2005learning,xian2020structure,qi2018geonet,liu2015deep,9316778, 8945171} are widely researched in the early stage with convolutional neural network (CNN) architectures.
Recently, the application of transformer model has become an active research topic~\cite{VIT,carion2020end,DPT,yang2021transformers, liu2024event,zhu2024scale}. 
For instance, DPT~\cite{DPT} utilizes vision transformers as the backbone instead of convolutional networks, resulting in globally coherent predictions. 
NeWCRFs~\cite{yuan2022newcrfs} employs fully-connected CRFs optimization within local windows to reduce the computation complexity. 
EReFormer~\cite{liu2024event} uses recurrent transformers to handle the challenges of sparse spatial information and rich temporal cues from event cameras.
\textcolor{black}{Though these methods have achieved promising performance on depth benchmarks, recent works have shifted focus towards the discrete regression mechanism due to its faster convergence.}
Discrete Regression based methods discretize the depth space to transform the regression problem into a classification problem~\cite{8764412, 9615228, 8010878,bhat2021adabins,li2022binsformer,10325550,cao2017estimating,huynh2020guiding,saha2021learning,lee2021depth}. 
For example, AdaBins~\cite{bhat2021adabins} adaptively generates depth division for each image and classify the pixels into them, which results in better performance and stronger generalization ability. 
\textcolor{black}{
Motivated by discrete regression based methods, we design a discrete regression based framework to model the plane bases and then weight the plane bases to predict per-pixel plane coefficients.
With this design, we achieve high efficiency in plane prediction, boosting overall depth estimation in a coarse-to-fine manner.
}
\vspace{-0.5em}
\input{framework}
\subsection{Geometric Constraints for Depth Estimation}
\textcolor{black}{
Instead of enhancing the structure of the network, an alternative research direction focuses on incorporating geometric constraints into the depth estimation.
Some methods ~\cite{li2021structdepth,zhang2020geolayout} aim to introduce plane assumptions in specific scenes, enabling the network to learn the priors of the scene and construct effective plane constraints. 
}
For instance, Structdepth~\cite{li2021structdepth} adopts a Manhattan-world assumption in indoor settings, ensuring that key surfaces (floor, ceiling, and walls) align with dominant directions. 
\textcolor{black}{
Some other methods~\cite{yu2020p,liu2018planenet,yang2018recovering,yin2019enforcing} design additional loss functions to constrain the 3D structure of the scenes.
For example, VNL~\cite{yin2019enforcing} proposes a novel virtual normal constraint, which disentangles the scale information and enriches the model with better shape information.
}
Most recently, some methods~\cite{long2021adaptive,yu2019single} explore to utilize plane prediction to guide the monocular depth estimation.
For instance, P3Depth~\cite{patil2022p3depth} regresses the plane coefficients of pixels and directly predicts the coplanarity relationship between pixels.
NDDepth~\cite{shao2023nddepth} introduces an extra prediction head to directly predict planes corresponding to pixels.
\textcolor{black}{
In contrast, our approach initially predicts a set of discrete plane bases and combines these bases to obtain pixel-wise plane coefficients. 
Finally, our method efficiently aggregates features related to planes and accurately predicts pixel-wise depth with geometric constraints.
}

\section{Methodology}\label{method}
This section describes the proposed method in detail. Figure~\ref{fig:2} illustrates the overall architecture of our network.
\subsection{Overview}

Given an input image $I$, the monocular depth estimation aims to predict the corresponding dense depth map $D$.
As show in Figure~\ref{fig:2}, we apply image encoder~\cite{he2016deep,liu2021Swin,radford2021learning} to extract the multi-scale image features.
First, we initialize a set of plane queries to interact with the image features and implicitly encode the plane information.
For the update of plane queries, we propose the adaptive plane query aggregation (APQA) module to aggregate image features at each scale, ensuring the consistency among the predicted hierarchical plane queries.
Then the updated plane queries are passed into the plane guided depth generator (PGDG) to predict adaptive ``plane bases" for each scene and the probabilities of pixels belonging to each plane basis. 
As a result, the plane coefficients corresponding to each pixel can be calculated through weighting the plane bases by probabilities.
Finally, we use camera intrinsic to transform the per-pixel plane coefficients into depth.
\subsection{Adaptive Plane Query Aggregation Module}
\textcolor{black}{To aggregate image features of different scales, we incorporates a novel adaptive plane query aggregation module.
Given the extracted multi-scale image features $\{F_l\}_{l=0}^{3}$, we integrate them into the plane queries $P \in \mathbb{R}^{L\times D}$ hierarchically. 
The previous work Mask2Former~\cite{cheng2022masked} takes the image features as key and value, and utilizes object queries to progressively aggregate multi-scale image features at each layer, which has been widely proven to be an efficient feature aggregation approach.
However, when a similar structure is introduced to aggregate plane information, we observe inconsistent activation regions for the same plane query when aggregating features from different scales, as shown in the first row of Figure~\ref{fig:inconsistency}. This way of updating only the queries leads to the model erroneously aggregating features from unrelated plane regions.
To address this issue, we design Adaptive Feature Modulator (AF Modulator), which modulates features with the plane queries from the previous layer and enables the top-down propagation of learned plane information.
Instead of taking image features as key and value, the AF Modulator takes image features as query and makes it interact with the plane queries at each scale. 
This approach effectively modulates the image features to align with those of previous layers and maintains consistent focus of the plane query on the same plane regions during feature interactions.
As shown in the second row of Figure~\ref{fig:inconsistency}, after feature modulation, our method ensures consistent activation regions for the same query across different layers, enabling more accurate and efficient feature aggregation in plane-related regions.
\begin{figure}[t]
  \centering 
  \includegraphics[width=\linewidth]{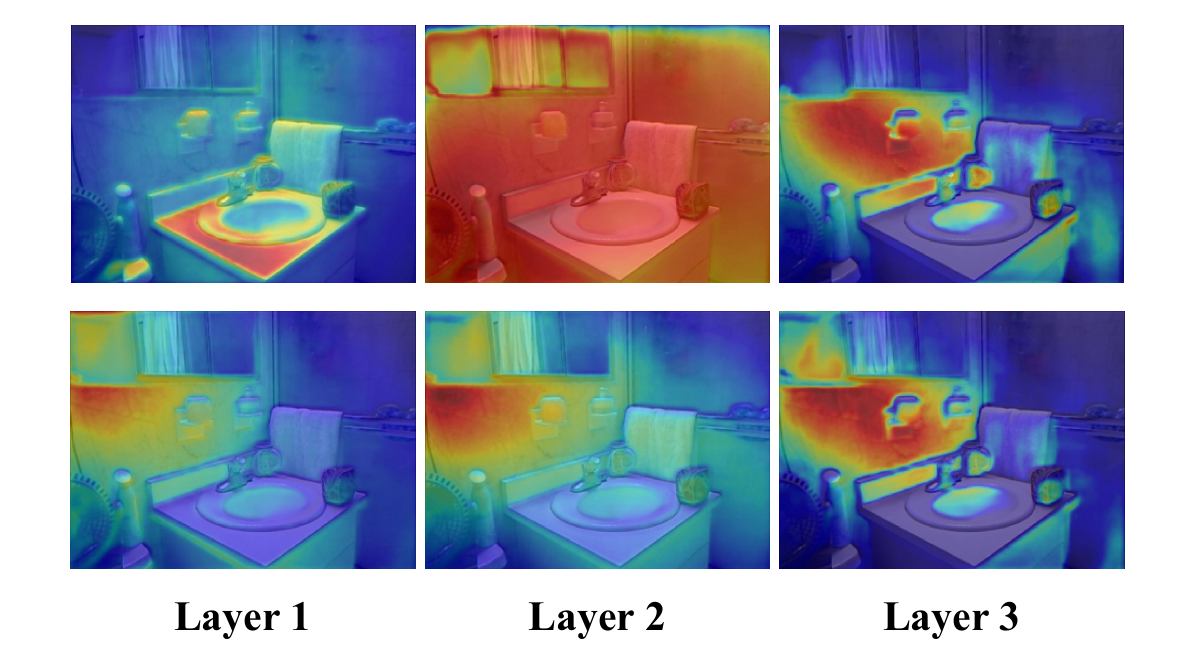} 
  \caption{\textbf{The visualization of query activation maps between plane queries and image features.}  \textbf{Top}: The query activation maps in Mask2former~\cite{cheng2022masked}. \textbf{Bottom}: Our query activation maps. The red regions indicate high correlation, while the blue regions indicate low correlation. Our plane queries can adaptively aggregate plane features in the image and predict plane bases in the scene. Each plane query focuses on distinct plane regions in the scene. } 
  \label{fig:inconsistency} 
\end{figure}
}

\textbf{Masked Transformer Layer.}
\textcolor{black}
{To facilitate the interaction between queries and image features, we utilize the masked attention in transformer as: 
\begin{equation}
    {P}_{l} =  \operatorname{softmax}(\mathcal{M}_{l-1} + Q_l K_l^{\top})V_l + {P}_{l-1},  \label{(1)}
\end{equation}
the attention mask $\mathcal{M}_{l-1}$ is defined as:
\begin{equation}
\mathcal{M}(x,y) = 
\begin{cases}
    0, & \text{if } M_{l-1}(x,y) = 1 \\
    -\infty, & \text{if } M_{l-1}(x,y) = 0 \\
\end{cases}
\end{equation}
where $l$ is the layer index, ${P}_{l-1}$ is the updated plane queries from the last layer, $Q_l$ is the query projection from ${P}_{l-1}$, $K_l$ and $V_l$ are the key and value projection of image features $F_l$, and $M_{l-1}$ is the binary attention mask predicted by the last layer.
}
In the cross-attention layer, regions with a mask value of 0 are excluded from the calculation, which can focus attention on localized features to accelerate convergence and improve overall efficiency.
\textcolor{black}{
Each masked transformer layer outputs a set of plane queries, which will be sent to the plane guided depth generator to obtain the plane coefficients per pixel and transform to depth.
}

\textbf{Adaptive Feature Modulator.}
\textcolor{black}{To modulate features to adapt to the plane queries of the previous layer, we introduce an additional transformer layer at each scale} (AF modulator in Figure \ref{fig:2}).
Specifically, we flatten the image features as query and use the plane queries as key and value in the transformer layer as:
\textcolor{black}{
\begin{equation}
    \hat{F}_l =  \operatorname{softmax}({\bar{Q}_l \bar{K}_l^{\top}}) \bar{V}_l + F_l,
\end{equation}
where $\bar{Q}_l$ is the projection of image features $F_l$, and $\bar{K}_l, \bar{V}_l$ are the projection of the plane queries $P_{l-1}$.
}
Then we can use $\hat{F}_i$ as the updated image features for subsequent interactions.
\textcolor{black}{The updated image features have a stronger correlation with the plane queries of the previous layer, emphasizing the guidance of high-level features. Combined with this design, we adaptively implement top-down multi-scale feature aggregation, thereby outputting consistent plane queries to predict pixel-by-pixel corresponding planes.}

\subsection{Plane Guided Depth Generator}
Motivated by Binsformer~\cite{li2022binsformer}, we apply a discrete regression manner to predict the plane coefficients via our plane-guided depth generator. 
In contrast to the previous methods, which directly use convolution networks to regress plane coefficients, we leverage plane queries to predict a set of discrete plane bases. 
\textcolor{black}{
Subsequently, the probability of pixels belonging to each plane basis can be predicted by calculating the similarity of plane queries and image features.
Consequently, the per-pixel plane coefficients can be calculated through weighted sum and finally transformed into depth maps by the 3D projection.
The proposed plane guided depth generator explicitly predict the planes corresponding to pixels and effectively utilize planes to guide depth estimation.
}

\begin{figure}[t]
  \centering 
  \includegraphics[width=0.8\linewidth]{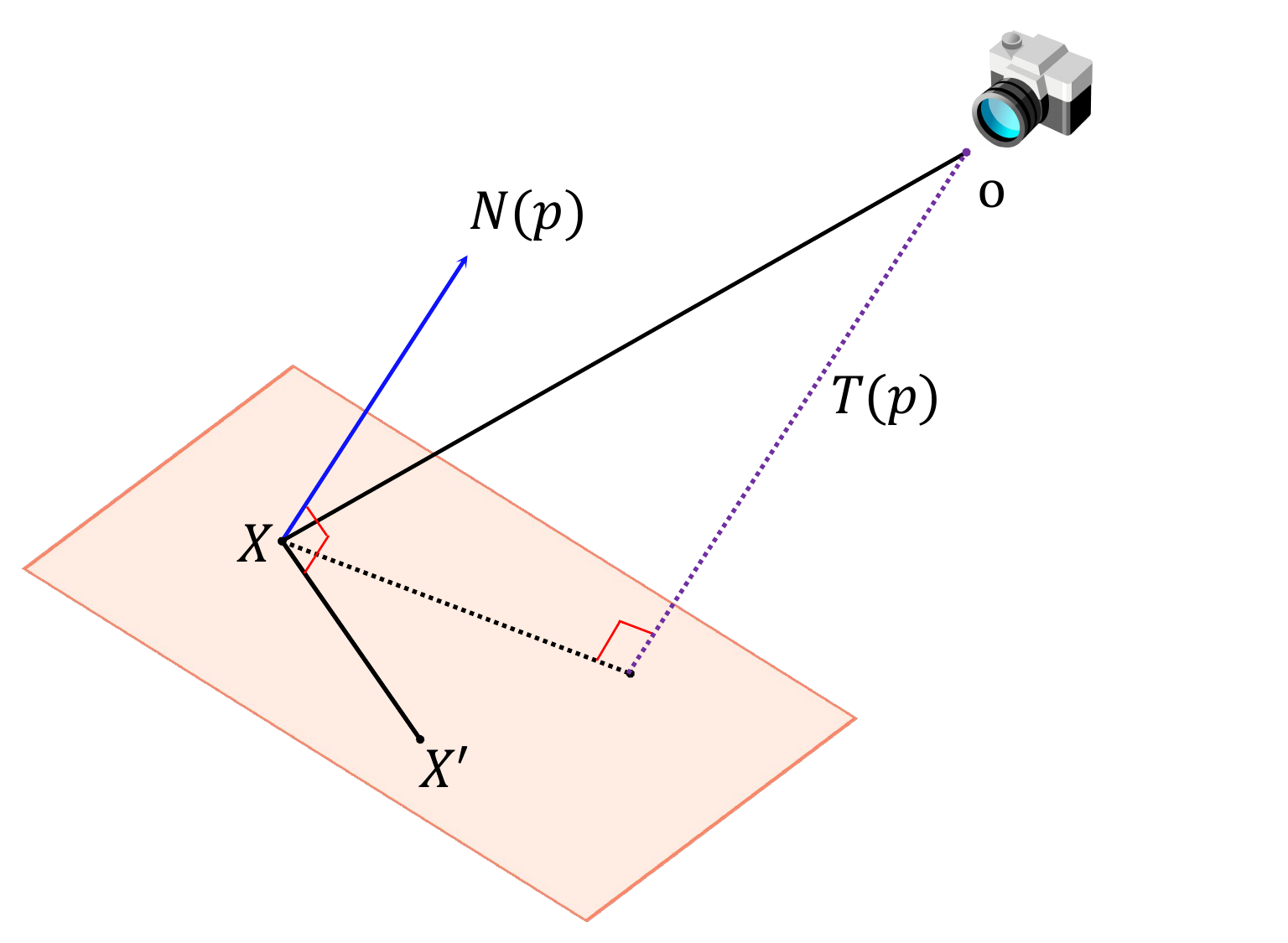} 
  \caption{\textbf{Illustration of the point-normal form equation.} $o$ represents the camera center, $N(p)$ represents the surface normal vector starting from $X$, $X{'}$ can be an arbitrary point on the plane, and $T(p)$ denotes the distance from the camera to the plane.} 
  \label{illustration} 
\end{figure}

\textbf{Plane Coefficient to Depth.}
To explicitly model planes in the scenes, we follow NDDepth~\cite{shao2023nddepth} to decompose 3D scenes into piece-wise planes and represent these planes using both the surface normal and the distance from the plane to camera center. 
Specifically, we use $X = (x, y, z)^{\top}$ to represent a 3D point in space, $p = (u, v)^{\top}$ to represent a 2D pixel when projecting $X$ to the image plane. 
The point-normal form equation uses a point on the plane and the surface normal of the plane to determine a plane in a 3D space. Specifically, the point-normal form equation can be formulated as:
\begin{equation}
    N(p)^{\top} (X{'} - X) = 0,
\end{equation}
where the surface normal $N(p) = (n_{1}, n_{2}, n_{3})^{\top}$ is the vector starting from $X$, $X{'}$ is a certain point on the plane and $o = (0, 0, 0)^{\top}$ is the camera center. Meanwhile, the distance from the plane to the camera center is:
\begin{equation}
T(p) = N(p)^{\top}X.
\label{eqa5}
\end{equation}
Figure~\ref{illustration} provides a more intuitive explanation of the above process.
Furthermore, according to the imaging principle of a pinhole camera, the projection of the 3D point $X$ to the 2D point $p$ using the camera intrinsic matrix $K$ can be formulated as:
\begin{equation}
    D(p) \tilde{p} = KX,
    \label{eqa6}
\end{equation}
where $\tilde{p}$ signifies the homogeneous coordinate of $p$, and $D(p)$ denotes the depth value of the point $p$.
\textcolor{black}{
Combining equation~\eqref{eqa5} and~\eqref{eqa6}, we can use the surface normal $N(p)$ and the distance $T(p)$ to calculate the depth value of the point $p$ as:
\begin{equation}
    D(p) = \dfrac{T(p)}{N(p)^{\top}K^{-1}\tilde{p}}. \label{eq:normal2depth}
\end{equation}
In this way, we enable the transformation from plane coefficients to depth values.
Compared with directly regressing depth values, our approach focus on the geometric correlations between pixels, which will be proven to be a simple and efficient way for depth estimation in our experiments.
}

\textbf{Normal and Distance Prediction.} 
\textcolor{black}{
Given the output plane queries $P \in \mathbb{R}^{L\times D}$, we pass them through independent MLPs to predict the plane bases, including surface normal $N \in \mathbb{R}^{L \times 3}$, distance $T \in \mathbb{R}^{L \times 1}$, and plane features $E \in \mathbb{R}^{L \times C}$, respectively.
}
\textcolor{black}{
To calculate the normal vector and distance for each pixel, we measure the similarity between plane features $E$ and image features $F$ to obtain probabilities as follows:
\begin{equation}
    \mathcal{S} = \operatorname{softmax}(E \times F^{\top}),
\end{equation}
where $F \in \mathbb{R}^{HW \times C}$ are the flatten features extracted by the image encoder. 
}
$\mathcal{S} \in \mathbb{R}^{L \times HW}$ represents the probabilities of pixels belonging to $L$ plane bases, which are used to weight the coefficients of the plane bases.
Therefore, we can calculate the surface normal map as:
\begin{equation}
    \mathcal{N} = N^{\top} \times \mathcal{S},
\end{equation}
and the distance map is obtained in the same way:
\begin{equation}
    \mathcal{T} = T^{\top} \times \mathcal{S},
\end{equation}
where $\mathcal{N} \in \mathbb{R}^{3 \times HW}$ and $\mathcal{T} \in \mathbb{R}^{1 \times HW}$ represent the predicted plane coefficients corresponding to pixels in the image.
The final depth is then calculated by Eq.~\eqref{eq:normal2depth}.
\textcolor{black}{
The normal and distance prediction first generates the plane basis, then predicts the probabilities of pixels belonging to these plane basis.
With the design of softly weighted sum, we enable the training of the network end-to-end, and alleviate the difficulty of directly regressing plane coefficients pixel by pixel.
}

It is worth noting that our method does not rely heavily on planar prior/assumption like previous methods (BTS~\cite{lee2019big}, P3Depth~\cite{P3Depth}), we equip the model with planar perception capabilities, while also maintaining strong performance in non-planar regions. 
Specifically, in non-planar regions, different pixels have different tangent planes. Our network predicts the coefficients of the tangent plane for each pixel through soft modeling, resulting in varying depths. Therefore, our network also performs well in scenes with few planes. Table~\ref{tab:kitti} demonstrates the effectiveness of our method in outdoor scenes with limited planes. P3Depth~\cite{P3Depth} relies on a piecewise planarity prior, resulting in a significant performance drop on the outdoor KITTI dataset compared to its strong performance on the indoor NYU-Depth-v2 dataset. In contrast, our method not only enhances performance on the NYU-Depth-v2 dataset but also achieves competitive results on the KITTI dataset without requiring special handling for non-planar regions.

\subsection{Loss Function}
To establish effective constraints on predicted depth, we use scale-invariant (SI) loss as described by Eigen~\cite{eigen2014depth}. 
Furthermore, we employ a loss based on cosine similarity to supervise the predicted normals, and a L1 loss to supervise the predicted distance, respectively.

\textbf{Depth Loss.} We use the SI loss to provide depth supervision. We first calculate the logarithm difference between the predicted depth map and the ground-truth depth as follows:
\begin{equation}
    \Delta D(p) = \log D(p) - \log D^{gt}(p),
\end{equation}
where $D(p)$ and $D^{gt}(p)$ are the predicted depth and GT depth value at pixel $p$, respectively. For K pixels with valid depth in an image, the SI loss is computed as:
\begin{equation}
    \mathcal{L}_D = \sqrt{\frac{1}{K}\sum_p{\Delta D(p)^{2}} - \frac{\lambda}{K^{2}}(\sum_p\Delta D(p))^{2}},
\end{equation}
where $\lambda$ is a variance minimizing factor. We set $\lambda$ to 0.15 in all experiments as customary.

\textbf{Plane Coefficient Loss.}
To effectively constrain the predicted plane coefficients, we use cosine similarity loss to supervise normal as:
\begin{equation}
    \mathcal{L}_\mathcal{N} = \frac{1}{K} \sum_p {1 - \mathcal{N}(p) {\mathcal{N}^{gt}}^{\top}(p)},
\end{equation}
where $\mathcal{N}$ and $\mathcal{N}^{gt}$ the predicted surface normal and GT surface normal.
And we adopt L1 loss to supervise distance as:
\begin{equation}
    \mathcal{L}_\mathcal{T} = \frac{1}{K} \sum_p |\mathcal{T}(p) - \mathcal{T}^{gt}(p)|,
\end{equation}
where $\mathcal{T}$ and $\mathcal{T}^{gt}$ the predicted plane-to-origin distance and GT distance.
Due to the absence of corresponding normals and distances ground truth in the NYU-Depth-v2 and KITTI datasets, we follow the approach from \cite{qiu2019deeplidar,shao2023nddepth} to derive the corresponding ground truth.
Finally, the total loss is defined as:
\begin{equation}
    \mathcal{L}_{total} = \alpha \mathcal{L}_D + \beta \mathcal{L}_\mathcal{N} + \gamma \mathcal{L}_\mathcal{T},
\end{equation}
where $\alpha$, $\beta$, and $\gamma$ are set to 10, 5, 1, respectively.

\section{Experiments}
\subsection{Datasets}
\noindent\textbf{NYU-Depth-v2}~\cite{Silberman:ECCV12} is an indoor dataset with $120K$ RGB-D videos captured from 464 indoor scenes of size $640 \times 480$. We use the official training/testing split to evaluate our method, where 249 scenes are used for training and 654 images from 215 scenes are used for testing. The max depth of the dataset is 10 meter.

\noindent\textbf{KITTI}~\cite{KITTI} is a widely used benchmark with outdoor scenes captured from a moving vehicle. There are two mainly used splits for monocular depth estimation. We employ the training/testing split proposed by Eigen~\cite{eigen2014depth} with 23488 training image pairs and 697 test images. And we set the maximum depth to 80 meter.

\noindent\textbf{SUN RGB-D}~\cite{song2015sun} is an indoor dataset captured by four different sensors and contains 10,000 RGB-D images. We perform generalization experiments on this dataset on the official test set of 5050 images with our model pretrained on NYUv2 dataset without fine-tuning.

\begin{figure*}[t]
  \centering 
  \includegraphics[width=\textwidth]{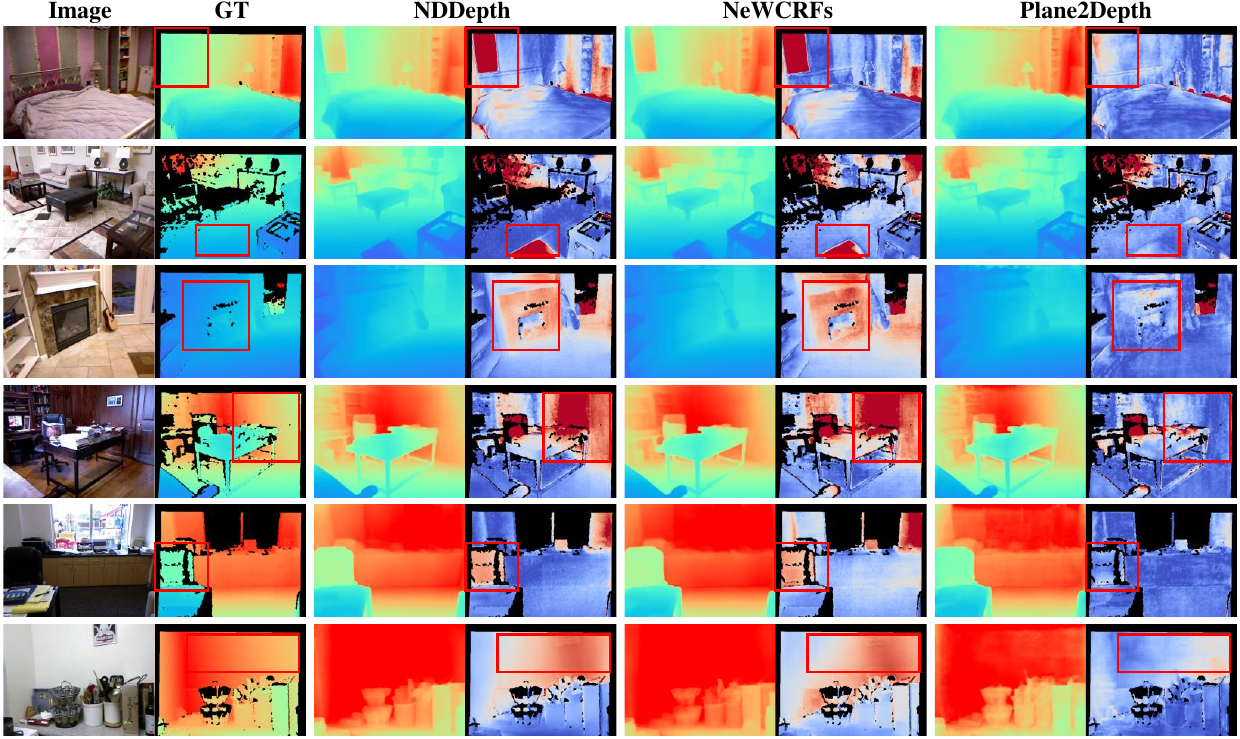} 
  \caption{\textbf{Qualitative results on NYU-Depth-v2 dataset.} Each column corresponds to a method. The predicted depth map is on the left and the error map is on the right. We apply the coolwarm colormap to visualize the error map, with values clipped at 0.5. Blue indicates low error, while red signifies high error. In row 1-3, we observe a clear advantage in our predicted maps for repetitive regions. Similarly, in row 4-6, our method effectively addresses challenges in areas with weak texture.} 
  \label{fig:3} 
  \vspace{-0.5em}
\end{figure*}

\input{nyutable}

\begin{figure*}[t]
  \centering 
  \includegraphics[width=\linewidth]{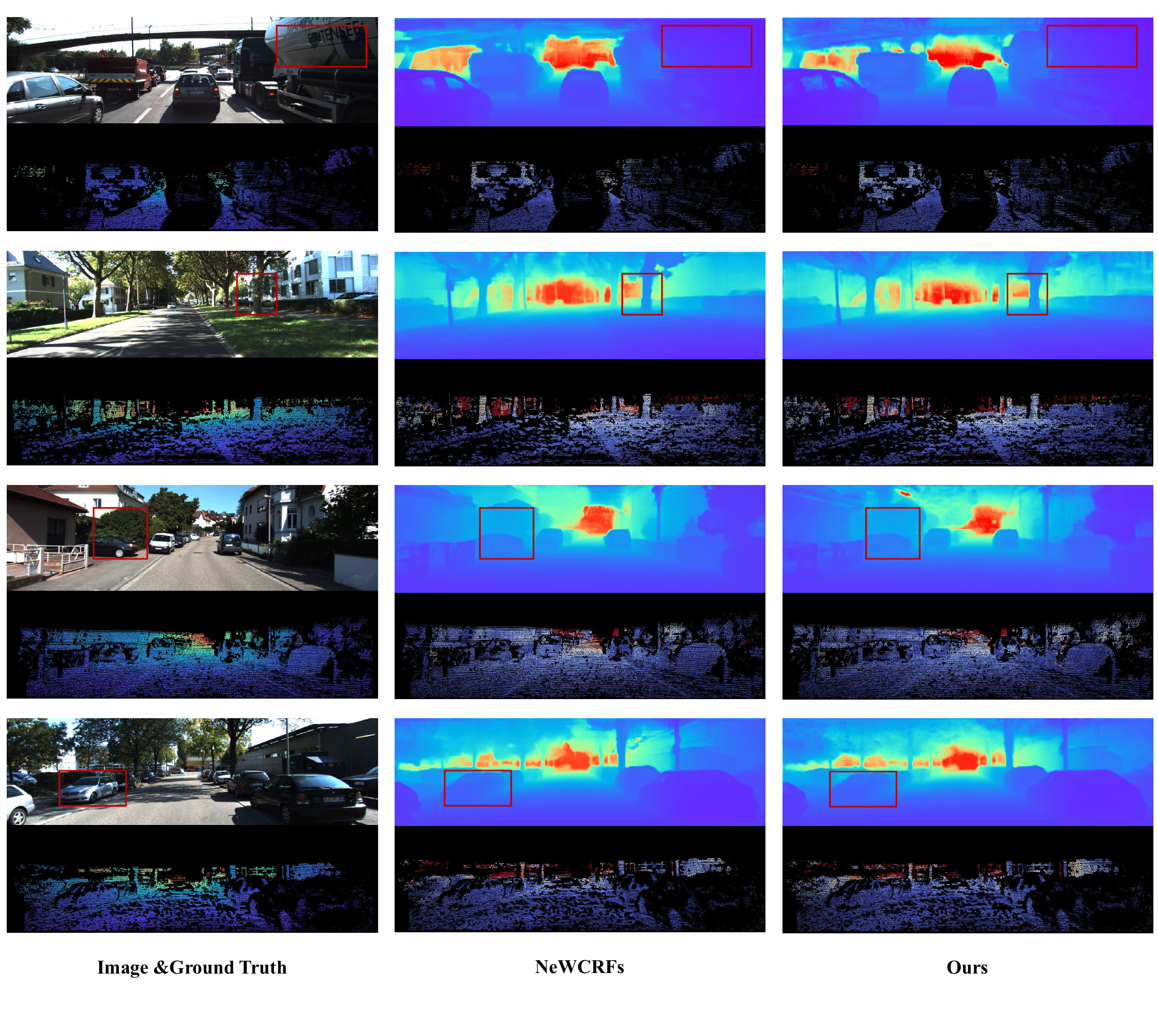} 
  \vspace{-3em}
  \caption{\textbf{The qualitative comparison on the KITTI dataset.} The first column presents the original image along with its corresponding depth ground truth. The subsequent columns illustrate the predicted depth maps and error maps generated by each model. We use color map \textit{rainbow} and \textit{coolwarm} to map the depth and error values respectively, and the error values are mapped using a unified range of 0-5m. Blue corresponds to lower (depth or error) values and red to higher values. The red box on the image represents areas of high curvature (such as the surfaces on the car, cylindrical tree trunks, leaves, etc.).
  Our approach yields satisfactory results in outdoor scenes thanks to the smooth modeling of planes, particularly in prominent plane areas.} 
  \label{fig:kitti_result} 
\end{figure*}
\input{kitti}

\begin{figure}[t]
  \centering 
  \includegraphics[width=\linewidth]{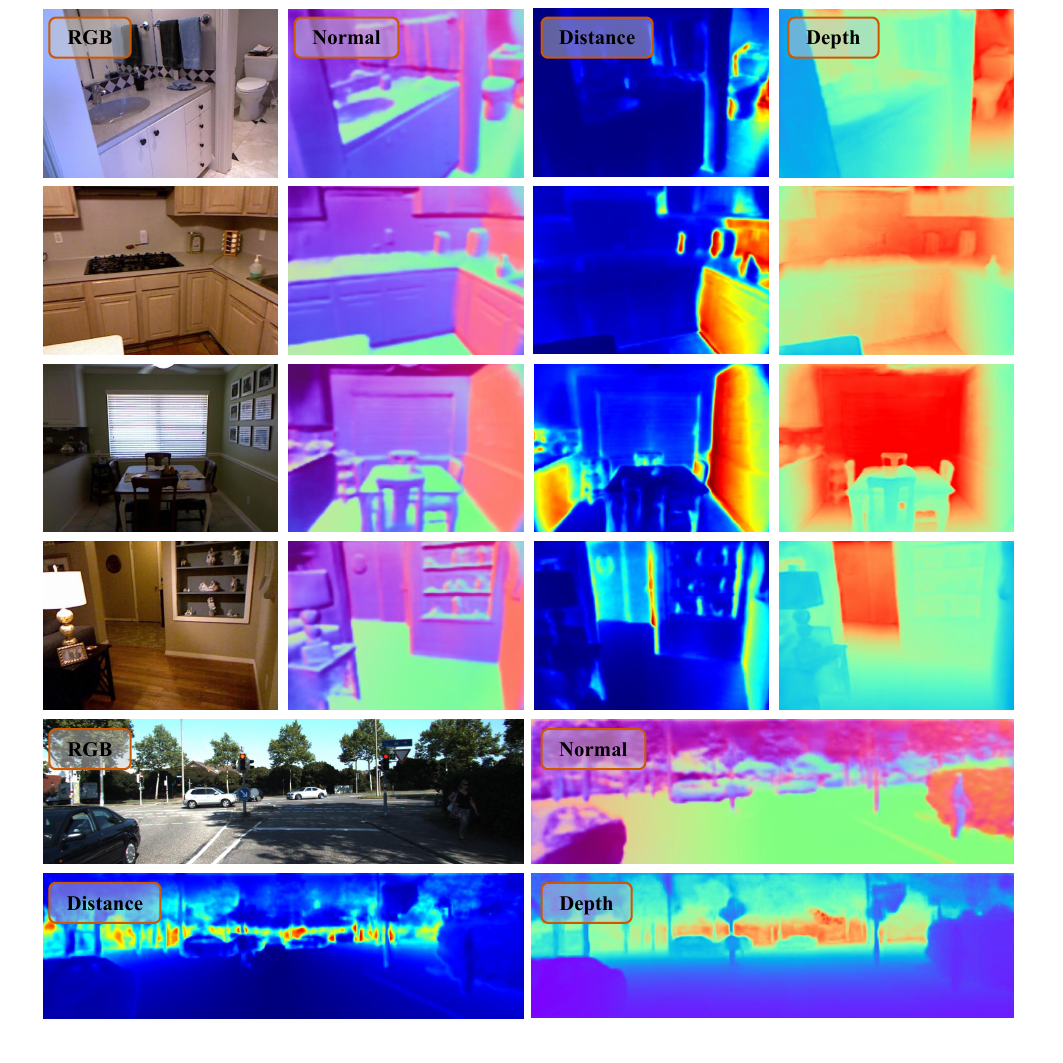} 
  \vspace{-2em}
  \caption{Visualization of the predicted surface normal, distance (from plane to camera center), and depth map on indoor and outdoor scenes. } 
  \label{fig:normal} 
\end{figure}

\subsection{Implementation Details}
Our work is implemented in PyTorch and experimented on 4 NVIDIA RTX 3090 GPUs. The network is optimized end-to-end with the Adam optimizer $(\beta_{1} = 0.9, \beta_{1} = 0.999)$. We set the default number of plane queries to 64. During training, we maintain the original resolution input for the NYU-Depth-v2 dataset and crop images from the KITTI dataset to a resolution of 352 $\times$ 1120.
The training runs for 40000 iterations with an initial learning rate of $1 \times 10^{-4}$ and batch size of $16$. 

\subsection{Network Details}
We randomly initialize plane queries using a set of learnable network parameters.
In practice, we use \textit{torch.nn.Embedding} to initialize plane queries with a default shape of $100 \times 256$. 
We employ the image encoder to extract multi-scale features with scale of $\frac{1}{4}, \frac{1}{8}, \frac{1}{16}, \frac{1}{32}$.
We then align the channel in preparation for subsequent interactions.
The features from the first three layers are utilized for interaction with plane queries, while the features from the final layer are employed to calculate the similarity between image pixels and plane features. 
In the decoder, we use MLP to predict normal and distance, respectively. After obtaining the normal and distance for each pixel, we normalize the normal vectors to ensure their unit length. The distance is then passed through a sigmoid function and multiplied by the maximum depth.

\subsection{Evaluation Metrics}
For evaluation, we use the error metrics used in previous work: the absolute relative error (AbsRel), squared mean relative error (SqRel), root mean squared error (RMSE) and its logarithmic form (RMSE log). We also use the threshold accuracy metrics $\delta_n$, indicating the percentage of pixels that satisfy $\max(d_i/\hat{d_i} , \hat{d_i}/d_i) < 1.25^n$ for $n = 1, 2, 3$. 
\subsection{Comparison to the State-of-the-art}
To demonstrate the effectiveness of our method, we compare it with the state-of-the-art methods on NYU, KITTI, and SUN RGB-D datasets.

\textbf{Evaluation on NYU-Depth-v2 dataset.}
Both the quantitative and qualitative results demonstrate the effectiveness of our method, particularly in areas with weak textures and repetitive patterns. 
Tabel~\ref{tab:nyu} presents the comparison of results between our method and other approaches. When using ResNet-50 as the backbone, our method outperforms P3Depth~\cite{patil2022p3depth} , which also employs a planar approach and uses ResNet-101 as the backbone. Additionally, with the Swin-Large backbone, our method outperforms all state-of-the-art methods on the NYU-Depth-v2 dataset.
%
%
In Figure~\ref{fig:3}, we conduct a qualitative comparison by visualizing depth maps and error maps. 
Specifically, we compare our method with NeWCRFs~\cite{yuan2022newcrfs} and NDDepth~\cite{shao2023nddepth}, revealing that our approach consistently produces more precise and accurate depth maps.
More importantly, our approach improves the depth prediction in areas with repetitive patterns (rows 1-3) and weak textures (rows 4-6). 
Though NDDepth incorporates plane constraints, its direct regression of plane coefficients through convolution does not fundamentally address this issue. 
In comparison, our network is able to distinguish the pixels belonging to the same plane even with different colors, resulting in more accurate and consistent depth predictions.
Therefore, it can be observed that our approach, which models plane information in space using plane queries, proves to be an effective solution to mitigate this problem. 

\begin{figure}[ht]
  \centering 
  \includegraphics[width=\linewidth]{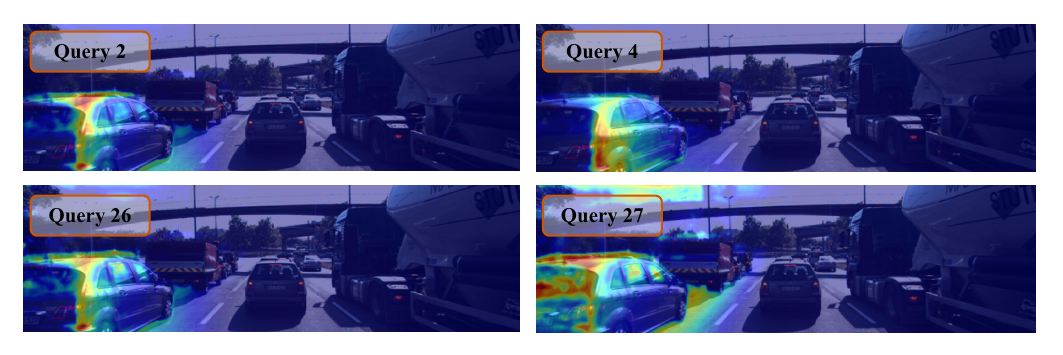}
\caption{\textbf{Attention map of different plane queries.} It can be observed that the curved surfaces of the car body are activated by multiple plane queries. This means that the plane coefficients for each pixel are influenced by multiple plane bases, allowing the coefficients to be adaptively calculated and resulting in accurate depth prediction.}
\label{fig:kitti_attmap}
\end{figure}

\textbf{Evaluation on KITTI dataset.}
As shown in Table \ref{tab:kitti}, we perform quantitative evaluations on the KITTI dataset.
Due to the large depth range (with a maximum depth of 80 meters) of the KITTI dataset, distant parts of the scene are projected onto very small planes, posing a substantial challenge in modeling these planes accurately.
Moreover, the scenes of KITTI exhibit numerous regions with high curvature, introducing an additional challenge to our plane-based depth prediction.
However, the results indicate that our method still achieves competitive performance in outdoor settings. 
We also show the visualization KITTI dataset in Figure~\ref{fig:kitti_result}, despite the complexity of outdoor scenes with many high-curvature regions, our method achieves satisfactory results in the outdoor scene, especially in some prominent plane regions.
In fact, in outdoor scenes, non-planar regions frequently occur. We attribute this success to our soft modeling of planes and the comprehensive interaction with image features. 
For each non-planar pixel, the network will predict different tangent plane (plane coefficients), resulting in different depth.
As shown in Figure~\ref{fig:kitti_attmap}, the curved surfaces of the car body are activated by multiple plane queries.  For an arbitrary point on the surface, its plane coefficients are the linear combination of all these activated plane bases weighted by its similarity to the plane queries.
Therefore, even in curved surface areas, we can accurately estimate the plane coefficients of the tangent plane.
Table~\ref{tab:kitti} demonstrates the effectiveness of our method in outdoor scenes with limited planes. P3Depth~\cite{patil2022p3depth} relies on a piecewise planarity prior, resulting in a significant performance drop on the outdoor KITTI dataset compared to its strong performance on the indoor NYU-Depth-v2 dataset. In contrast, our method not only enhances performance on the NYU-Depth-v2 dataset but also achieves competitive results on the KITTI dataset without requiring special handling for non-planar regions.
To illustrate the effectiveness of our method in modeling the plane, we also visualize the surface normal, distance, and corresponding depths of some outdoor and indoor scenes in Figure~\ref{fig:normal}.
Our network demonstrates the ability to model corresponding planes and learn the tangent plane for each pixel in non-planar regions correctly. Therefore, we can predict an accurate depth both indoors and outdoors.

\begin{figure}[t]
  \centering 
  \includegraphics[width=\linewidth]{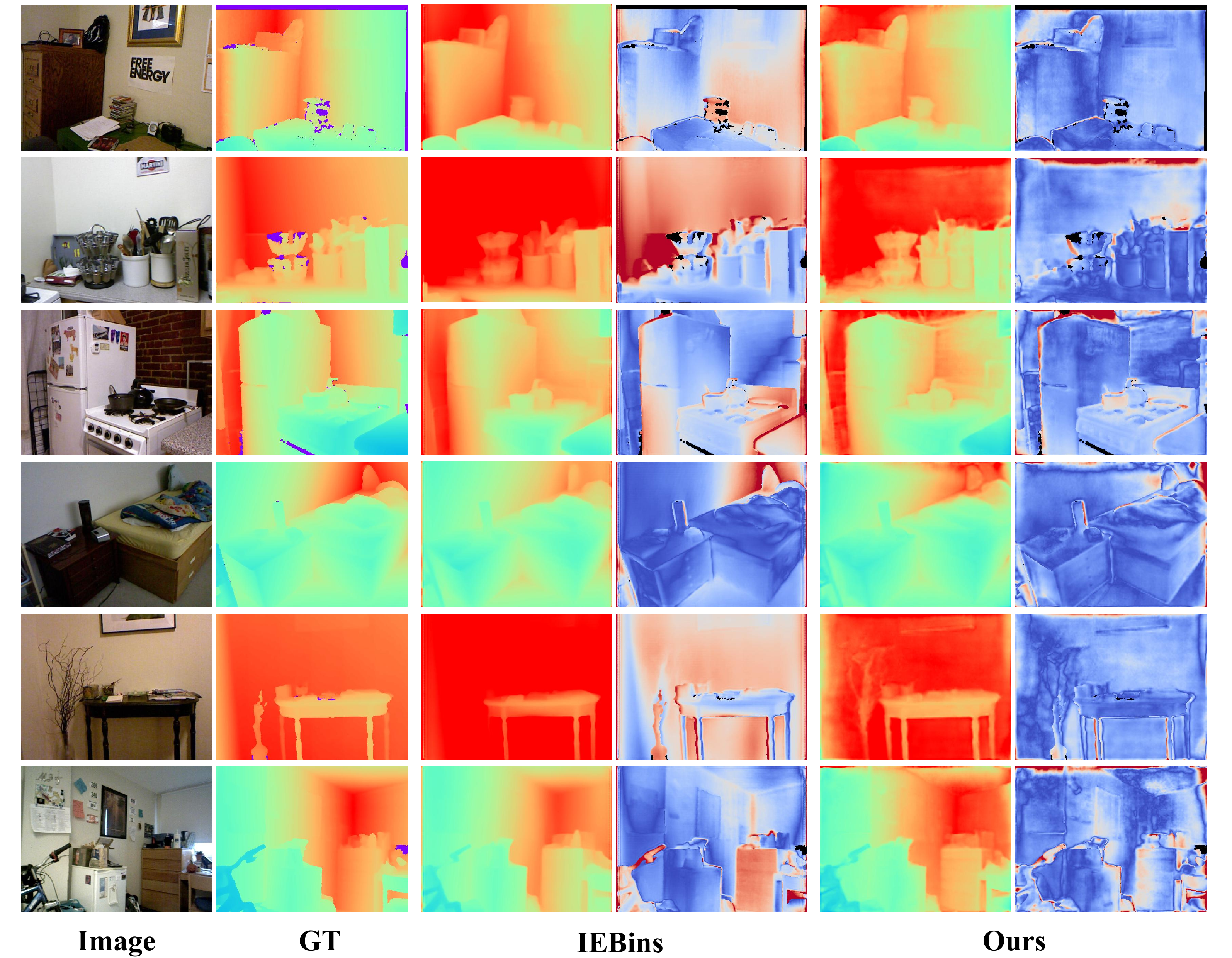} 
  \caption{\textbf{The qualitative zero-shot results on SUN RGB-D dataset.} The first two columns correspond to the input image and depth GT respectively. In the subsequent couples, the left ones correspond to the depth maps using color map \textit{rainbow}, and the right ones correspond to error maps using color map \textit{coolwarm}. We apply the coolwarm colormap to visualize the error map, with values clipped at 0.5. Blue indicates low error, while red signifies high error.} 
  \label{fig:sunrgbd} 
\end{figure}

\input{sunrgbd}

\textbf{Zero-shot Evaluation on SUN RGB-D dataset.}
To validate the generalization performance of our method, we conduct zero-shot testing on the SUN RGB-D dataset, and the results are shown in Table \ref{tab:3}. %
In particular, we use the model trained on NYU-Depth-v2 for the zero-shot evaluations. 
Also, we resize the images to match the training resolution during evaluation. 
The results indicate that our method outperforms previous state-of-the-art approaches, highlighting its robust generalization ability to effectively model planes in unseen scenarios.
Furthermore, we provide visualization results in Figure~\ref{fig:sunrgbd}, where our model demonstrates outstanding generalization performance, especially in scenes with a more regular structure. The error map reveals that our method achieves exceptional results in the planar areas of the scene, surpassing IEBins~\cite{shao2023iebins}. This suggests that our approach is capable of effectively modeling the planes of unseen scenes as well.

\textbf{Comparison with relative depth estimation methods.} As recent advancements, Depth Anything~\cite{depthanything} and GeoWizard~\cite{fu2024geowizard} have achieved remarkable generalization performance, and we also compare our method with them, as shown in Table~\ref{tab:relative}. Benefiting from a robust backbone and diverse training data, these relative depth estimation models exhibit impressive performance when generalized to the indoor NYUv2 dataset. However, due to the absence of outdoor data in the training set, their generalization to the KITTI dataset is relatively poor. Notably, during evaluation, these methods require ground truth (GT) from the test set to compute scale and shift, which are essential for converting relative depth to metric depth.

\subsection{Efficiency experiments.}
To verify the efficiency of our method, we compare the latency of our method with other baselines on the KITTI dataset. as shown in Table~\ref{tab:latency}. The results show that with the same backbone and image size, our method also achieves satisfactory performance in terms of efficiency. Thanks to our AF Modulator, we achieved better performance with fewer transformer layers, which also reduced inference time and improved efficiency.

\input{newtable}
\input{latency}
\subsection{Additional Visualization Results}
In Figure~\ref{fig:att_map}, we visualize the activation maps of the plane queries to represent the regions of interest in the image. The activation maps reveal that the activation areas for each query in the scene share a similar orientation, which means each plane query representing a specific orientation.
When dealing with larger planes in the scene, such as floors and walls, a single query may struggle to aggregate such a vast area.
As a result, multiple plane queries are employed to aggregate a larger plane, with each part of the plane aggregated by different query.
%
%
Moreover, we show the reconstructed point clouds on the NYU-Depth-v2 dataset in Figure~\ref{fig:point_cloud}. 
In comparison to NeWCRFs~\cite{yuan2022newcrfs}, our method produces 3D scenes with more uniform planes and comprehensive details. This underscores the exceptional advantage of our approach in perceiving the geometry of the scenes.

\input{fig}

\subsection{Ablation Study}
In this section, we study the effectiveness of the proposed module and the impact of different designs in our network.

\textbf{Adaptive Plane Query Aggregation Module.} 
To demonstrate the effectiveness of incorporating our APQA module, we conduct ablation experiments on the NYU-Depth V2 dataset. 
%
%
In our network, we apply a transformer layer to each level of features. Starting from the second layer, we use additional modulators on the features to ensure the plane consistency across multiple levels.
As shown in Table \ref{tab:4}, we compare the results of using 0, 3, 6, 9 layers without using our modulator module. Note that we bypass the whole APQA module for the abalation study of 0 layer.
As a result, our method achieves a 6.82\% improvement in RMSE compared to the 3-layer approach. 
Furthermore, despite increasing the number of transformer layers (from 3 to 5), the performance was even better than using 6 and 9 layers (with 2 and 3 layers for each scale of image features, respectively). 
With less transformer layers, our approach surpasses models with a greater number of layers, highlighting the significance of the adaptive feature modulator. 
The modulation of multi-scale image features guarantees the conveyance of global contextual information, thereby maintaining query consistency throughout the interaction process and yielding superior depth predictions.
In Figure~\ref{fig:att_map}, we visualize the query activation maps to represent the regions of interest in the image. The activation maps reveal that the plane queries adaptively predict the ``plane bases" in the scene, with each plane base representing a specific orientation.
This indicates the effectiveness of our plane queries.

\input{ablation2}

\input{ablation1}

\textbf{Plane Coefficient Constraints.} 
To validate the effectiveness of the constraints on the predicted surface normal and plane-to-origin distance, we conduct an ablation study on the NYU-Depth V2 datasets.
In the absence of a planar constraint, we employ a bin-based discrete regression approach~\cite{li2022binsformer} to predict depth, ensuring the coherence of the network.
As shown in Table~\ref{tab:5}, we compare the models with and without the constraints on normal and distance. 
Our results show that the performance is significantly improved after adding both the normal and distance constraints. 
This highlights the essential need to incorporate plane constraints and the effectiveness of our approach in modeling planes.
When we apply only a single normal or distance constraint, we observe a decline in the model's performance. 
We hypothesize that this is due to the imposition of a sole constraint leading to biased predictions in plane coefficients. 
Therefore, it is essential to jointly constrain both the normal vector and the distance for improved accuracy.

\input{ablation_qnum}

\textbf{The Number of Plane Queries.}
We explore the impact of varying the number of plane queries on performance in Table~\ref{tab:6}. For efficiency, we use a batch size of 8 in this experiment. It is observed that the number of 32 queries produces optimal performance. We posit that the scarcity of queries makes it challenging to encompass all the plane bases in the scene, whereas an excessive number of queries would add complexity to the network's learning process.

\begin{figure*}[t]
  \centering 
  \includegraphics[width=\linewidth]{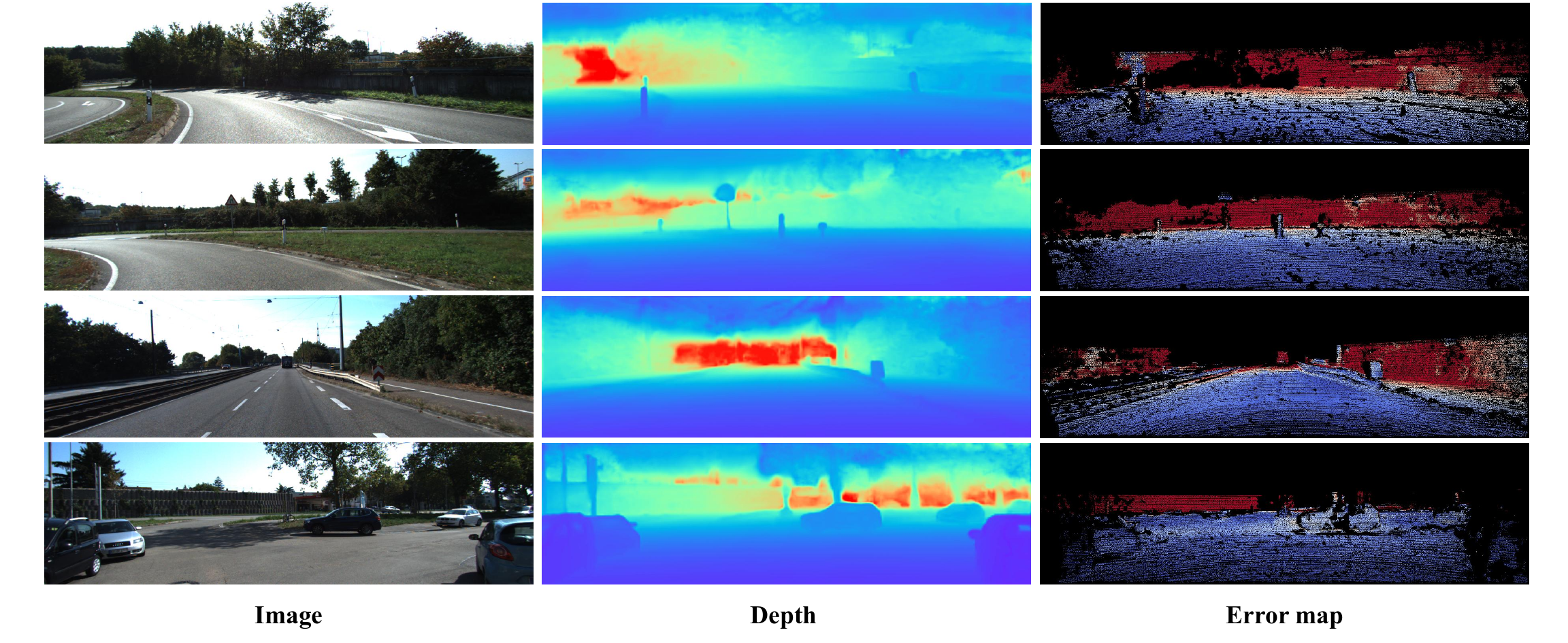} 
  \caption{\textbf{The failure cases in outdoor scenes.} We use color map \textit{coolwarm} to map the error values, the error values are mapped using a unified range of 0-5m. Blue corresponds to lower (depth or error) values and red to higher values.} 
  \label{fig:failurecase} 
\end{figure*}
\section{Limitations}
We observe that in certain scenarios, our method can produce relatively larger errors. As illustrated in Figure~\ref{fig:failurecase}, in outdoor scenes where both sides of the road consist of complex fine-grained surfaces such as leaves and lack large planar features for reference, the model's predicted plane coefficients may be inaccurate. Furthermore, in areas with many curved surfaces, the model's predictions are also affected, leading to less accurate depth estimation.
We believe this is due to the lack of planes with similar orientations to these surfaces. In an area full of curved surfaces, the plane coefficients of pixels vary, making it somewhat challenging for the network to predict these coefficients. In such cases, the presence of nearby planes with similar orientations (such as walls) can help the network better estimate the plane coefficients of the pixels on curved surfaces.

\section{Conclusions}
In this paper, we introduce a plane guided hierarchical adaptive framework for monocular depth estimation, which includes a plane guided depth generator with an adaptive plane query aggregation module. We use plane information modeled through queries to tackle the challenges of monocular depth estimation in areas with weak textures and repetitive patterns. 
Our approach excels not only in predicting precise and consistent depth maps but also in concurrently forecasting plane information within the scene.
Extensive experiments demonstrate the effectiveness of our approach, surpassing previous state-of-the-art methods. 
Thanks to our soft modeling of plane information, we can address high-curvature regions in the scene to some extent. However, as these areas are not explicitly considered, we believe that there is still ample room for exploration to further enhance performance. We consider this as part of our future work.

\bibliographystyle{IEEEtran}
\bibliography{reference}

\vfill

\end{document}

%% file: framework.tex
\begin{figure*}[!ht]
  \centering 
  \includegraphics[width=\linewidth]{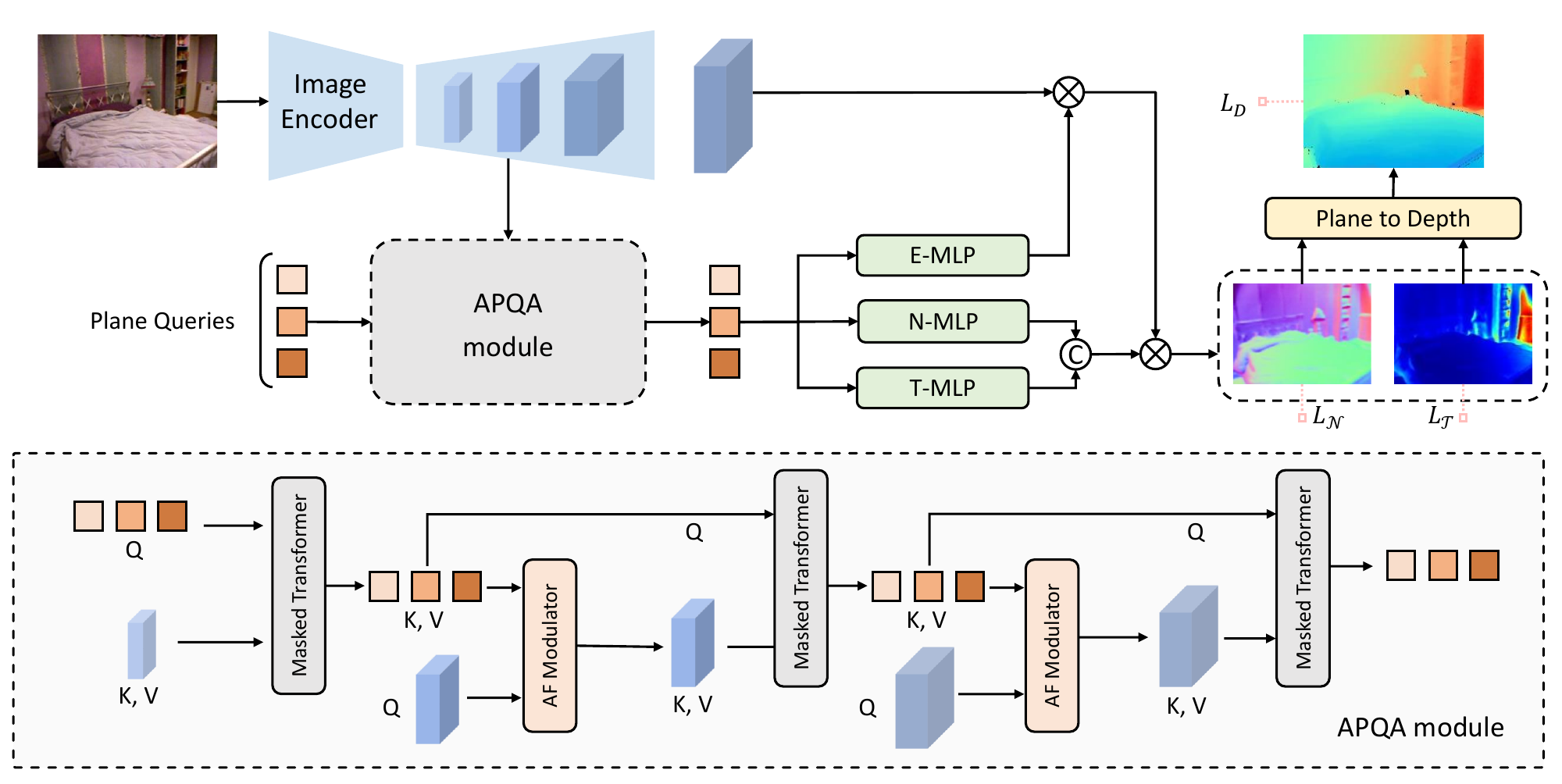} 
  \caption{\textbf{The overall architecture of Plane2Depth.} We use a set of plane queries to predict plane coefficients through E-MLP, N-MLP, and T-MLP, respectively. Then the predicted plane coefficients are converted to metric depth maps through the pinhole camera model. For consistent query prediction, we adopt the APQA module to aggregate multi-scale image features and adaptively modulate them via AF modulators. } 
  \label{fig:2} 
  \vspace{-0.5em}
\end{figure*}

%% file: nyutable.tex
\begin{table*}[ht]
  \centering
  \caption{Quantitative comparison on the NYU-Depth-v2 dataset. The best results are in \textbf{bold}, and the second best are \underline{underlined}. }
  \label{tab:nyu}%
    \begin{tabular}{c||c||ccc||ccc}
    \toprule
    Method & Backbone   & RMSE ↓ & Abs Rel ↓ &  log10 ↓  & $\delta < 1.25$ ↑& $\delta < 1.25^2$ ↑& $\delta < 1.25^3$ ↑ \\
    \midrule
    Fu et al.~\cite{DORN} \textit{(CVPR 2018)} & ResNet-101 & 0.509 & 0.115 & 0.051 & 0.828 & 0.965 & 0.992 \\
    VNL~\cite{yin2019enforcing} \textit{(CVPR 2019)} & ResNeXt-101  & 0.416 & 0.108 & 0.048 & 0.875 & 0.976 & 0.994 \\
    BTS~\cite{lee2019big} \textit{(arXiv 2019)} & DenseNet-161 & 0.407 & 0.113 & 0.049 & 0.871 & 0.977 & 0.995 \\
    Adabins~\cite{bhat2021adabins} \textit{(CVPR 2021)} & EfficientNet-B5+MiniViT & 0.364 & 0.103 & 0.044 & 0.903 & 0.984 & 0.997 \\
    P3Depth~\cite{patil2022p3depth} \textit{(CVPR 2022)} & ResNet-101 & 0.356 & 0.104 & 0.043  & 0.898 & 0.981 & 0.996 \\
    NeWCRFs~\cite{yuan2022newcrfs} \textit{(CVPR 2022)} & Swin-Large & 0.334 & 0.095 & 0.039 & 0.927 & 0.991 & \underline{0.998} \\
    BinsFormer~\cite{li2022binsformer} \textit{(TIP 2024)} & Swin-Large & 0.330 & 0.094 & 0.040 & 0.925 & 0.989 & 0.997 \\
    HA-Bins~\cite{10325550} \textit{(TCSVT 2023)} & Swin-Large & 0.334 & 0.094 & 0.040 & 0.922 & 0.991 & \underline{0.998} \\
    iDisc~\cite{piccinelli2023idisc} \textit{(CVPR 2023)} & Swin-Large & 0.313 & 0.086 & 0.037 & 0.936 & 0.992 & \underline{0.998} \\
    NDDepth~\cite{shao2023nddepth} \textit{(ICCV 2023)} & Swin-Large & 0.311 & 0.087 & 0.038 & 0.936 & 0.991 & \underline{0.998 }\\
    IEBins~\cite{shao2023iebins} \textit{(NeurIPS 2023)} & Swin-Large & 0.314 & 0.087 & 0.038 & 0.936 & 0.992 & \underline{0.998} \\
    Wordepth~\cite{zeng2024wordepth} \textit{(CVPR 2024)} & Swin-Large & 0.317 & 0.088 & 0.038 & 0.932 & 0.992 & \underline{0.998} \\
    \midrule
    & ResNet-50 & 0.350 & 0.101 & 0.042 & 0.899 & 0.983 & 0.997 \\
    \textbf{Plane2Depth(Ours)}
    & Swin-Large & \underline{0.308} & \underline{0.084} & \underline{0.036} & \underline{0.940} & \underline{0.993} & \underline{0.998} \\
     &Convnext-Large & \textbf{0.268} & \textbf{0.074} & \textbf{0.032} & \textbf{0.956} & \textbf{0.994} & \textbf{0.999} \\
    \bottomrule
    \end{tabular}%
\end{table*}%

%% file: kitti.tex
\begin{table*}[t]
\setlength{\tabcolsep}{4pt}
\centering
\caption{Quantitative depth comparison on the Eigen split of the KITTI dataset.}
\label{tab:kitti}%
\begin{tabular}{c||c||cccc||ccc}
\toprule
Method   &Backbone     & AbsRel ↓ & RMSE ↓  & RMSE log ↓ & SqRel ↓ & $\delta < 1.25$ ↑ & $\delta < 1.25^2$ ↑ & $\delta < 1.25^3$ ↑ \\ 
\midrule
VNL~\cite{yin2019enforcing} \textit{(CVPR 2019)}   &ResNeXt-101        & 0.072  & 3.258 & 0.117 &-  & 0.938 & 0.990 & 0.998 \\
BTS~\cite{lee2019big} \textit{(arXiv 2019)}    &DenseNet-161       & 0.060  & 2.798 & 0.096 & 0.249  & 0.955 & 0.993 & 0.998\\
Adabins~\cite{bhat2021adabins} \textit{(CVPR 2021)}   &EfficientNet-B5+MiniViT    & 0.060  & 2.372 & 0.090  & 0.197 & 0.963 & 0.995 & \textbf{0.999} \\
P3Depth~\cite{patil2022p3depth} \textit{(CVPR 2022)} & ResNet-101 & 0.071 & 2.842 & 0.103 & 0.270 & 0.953 & 0.993 & 0.998 \\
NeWCRFs~\cite{yuan2022newcrfs} \textit{(CVPR 2022)}   &Swin-Large    & 0.052  & 2.129 & 0.079  & 0.155 & 0.974 & \underline{0.997} & \textbf{0.999} \\
BinsFormer~\cite{li2022binsformer} \textit{(TIP 2024)}  &Swin-Large  & 0.052  & 2.098 & 0.079  & 0.151 & 0.974 & \underline{0.997} & \textbf{0.999} \\
iDisc~\cite{piccinelli2023idisc} \textit{(CVPR 2023)}   &Swin-Large    & \underline{0.050}  & 2.067 & 0.077 & 0.145  & 0.977 & \underline{0.997} & \textbf{0.999} \\ 
NDDepth~\cite{shao2023nddepth} \textit{(ICCV 2023)}  &Swin-Large     & \underline{0.050}  & \underline{2.025} & \underline{0.075} &\underline{0.141}   & \underline{0.978} & \textbf{0.998} & \textbf{0.999} \\
IEBins~\cite{shao2023iebins} \textit{(NeurIPS 2023)}  &Swin-Large     & \underline{0.050}  & \textbf{2.011} & \underline{0.075} &\underline{0.142}   & \underline{0.978} & \textbf{0.998} & \textbf{0.999} \\
Wordepth~\cite{zeng2024wordepth} \textit{(CVPR 2024)}   &Swin-Large    & \textbf{0.049}  & 2.039 & 0.074 & -  & \textbf{0.979} & \textbf{0.998} & \textbf{0.999} \\ 
\midrule
&ResNet-50    & 0.059  & 2.368 & 0.088 & 0.201 & 0.968 & 0.995 & \textbf{0.999}\\ 
\textbf{Plane2Depth(Ours) }     &Swin-Large     & 0.051  & 2.050 & 0.077 & 0.147 & 0.976 & \underline{0.997} & \textbf{0.999}\\ 

&Convnext-Large & \textbf{0.049}  & \underline{2.023} & \textbf{0.073} & \textbf{0.139} & \textbf{0.979} & \textbf{0.998} & \textbf{0.999}\\ 

\bottomrule
\end{tabular}
\end{table*}

%% file: sunrgbd.tex
\begin{table}[t]
\small
\setlength{\tabcolsep}{1.2mm}
\centering
\caption{Zero-shot evaluation on the SUN RGB-D dataset. We use the model trained on NYU-Depth-v2 for evaluation.}
\label{tab:3}%
\begin{tabular}{c|ccc|cc}
\toprule
Method        & AbsRel & RMSE  & $log_{10}$ & $\delta < 1.25$  & $\delta < 1.25^{2}$ \\ 
\midrule
VNL~\cite{yin2019enforcing}           & 0.183  & 0.541 & 0.082 & 0.696 & 0.912 \\
BTS~\cite{lee2019big}           & 0.172  & 0.515 & 0.075 & 0.740 & 0.933 \\
Adabins~\cite{bhat2021adabins}       & 0.159  & 0.476 & 0.068 & 0.771 & 0.944 \\
Binsformer~\cite{li2022binsformer} & 0.143  & 0.421 & 0.061 & 0.805 & 0.964 \\
TrapAttention~\cite{ning2023trap} & 0.141  & 0.414 & 0.061 & 0.806 & 0.964 \\
NDDepth~\cite{shao2023nddepth}       & \textbf{0.137}  & \underline{0.411} & \underline{0.060} & \underline{0.820} & \underline{0.970} \\
\midrule
Ours          & \underline{0.140}  & \textbf{0.409} & \textbf{0.058} & \textbf{0.826} & \textbf{0.972} \\
\bottomrule
\end{tabular}
\end{table}

%% file: newtable.tex
\begin{table*}[ht]
\centering
  \caption{Quantitative comparison with methods for relative depth estimation on the NYU-Depth-v2 and KITTI datasets. The best results are in \textbf{bold}. ``Data" refers to the number of training samples. $\dagger$ denotes methods that only predict relative depth.}
  \label{tab:relative}%
\begin{tabular}{@{}c|c|c|cc|cc@{}}
\toprule
\multirow{2}{*}{Method} & \multirow{2}{*}{Data} & \multirow{2}{*}{Backbone} & \multicolumn{2}{c|}{NYUv2} & \multicolumn{2}{c}{KITTI} \\
                        &                        &                           & AbsRel ↓       & $\delta < 1.25$ ↑          & AbsRel ↓      & $\delta < 1.25$ ↑         \\ \midrule
Depth Anything$^\dagger$~\cite{depthanything}          & 63.5M                  & DIVO V2                   & \textbf{0.043}        & \textbf{0.981}      & 0.076        & 0.947      \\
GeoWizard$^\dagger$~\cite{fu2024geowizard}               & 354K                   & Stable Diffusion V2       & 0.052        & 0.966       & 0.097        & 0.921      \\
\midrule
\textbf{Plane2Depth(Ours)}                    & 47K                    & Convnext-Large                & 0.074        & 0.956       & \textbf{0.049}        & \textbf{0.979}      \\ \bottomrule
\end{tabular}
\end{table*}

%% file: latency.tex
\begin{table}[t]
\centering

\caption{Comparison of the latency on KITTI. All methods  are evaluated on a single RTX 3090 GPU.}
\label{tab:latency}
\begin{tabular}{@{}c|c|l|l@{}}
\toprule
Method     & Backbone   & Image Size            & Time  \\ \midrule
NeWCRFs~\cite{yuan2022newcrfs}    & Swin-Large &   $352 \times 1216$                    & 150ms \\
Binsformer~\cite{li2022binsformer} & Swin-Large &   $352 \times 1216$                    & 270ms \\
NDDepth~\cite{shao2023nddepth}    & Swin-Large &   $352 \times 1216$                    & 250ms \\
iDisc~\cite{piccinelli2023idisc}    & Swin-Large &   $352 \times 1216$                    & 130ms \\
Ours       & Swin-Large & $352 \times 1216$ & 160ms \\ \bottomrule
\end{tabular}
\end{table}

%% file: fig.tex
\begin{figure*}[htbp]
  \centering 
  \includegraphics[width=0.9\linewidth]{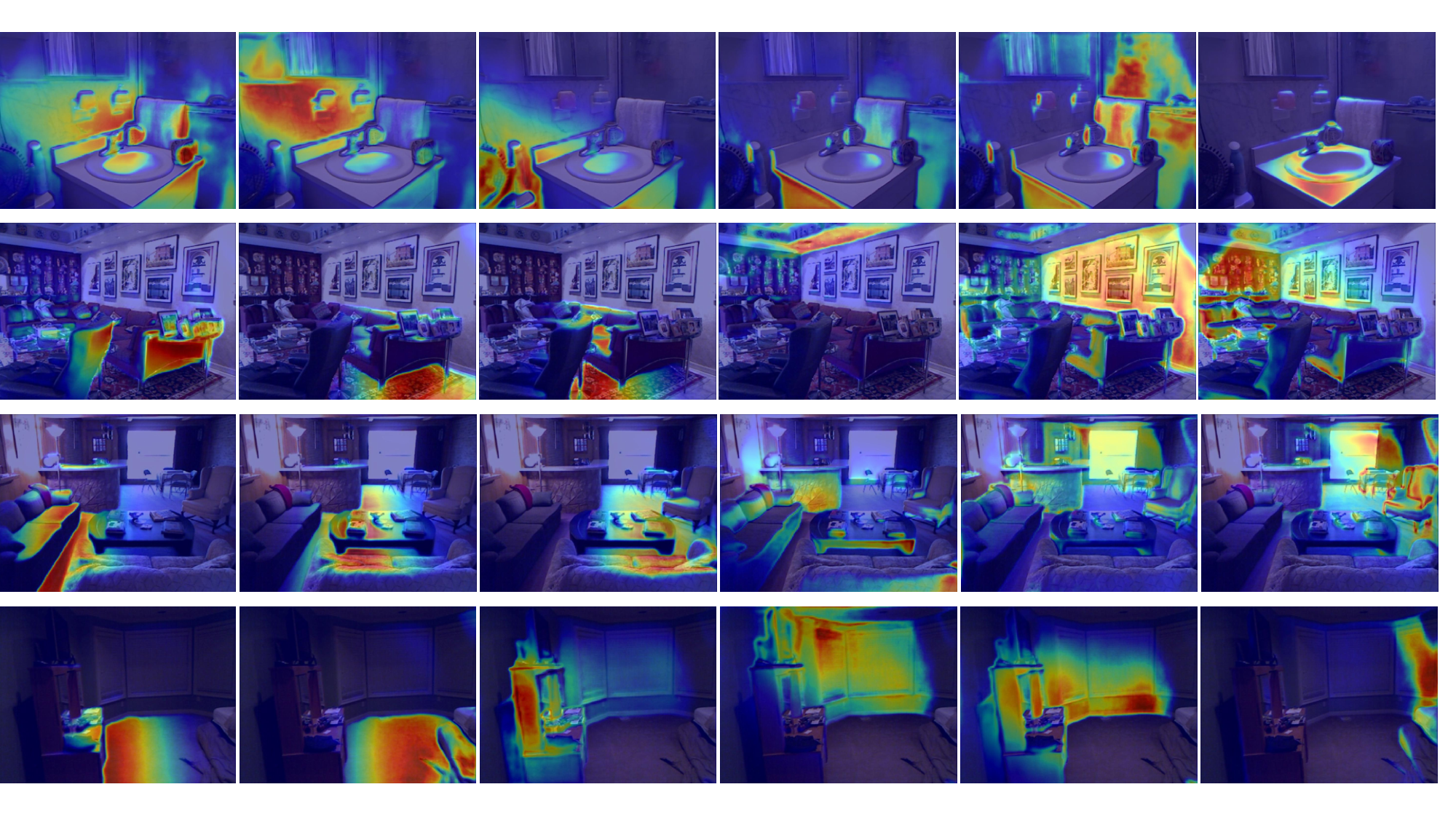} 
  \vspace{-1em}
  \caption{\textbf{The visualization of query activation maps between plane queries and image features.} The red regions indicate high correlation, while the blue regions indicate low correlation. Our plane queries can adaptively aggregate plane features in the image and predict plane bases in the scene. Each plane query focuses on distinct plane regions in the scene. } 
  \label{fig:att_map} 
  \vspace{3em}
  \centering 
  \includegraphics[width=\linewidth]{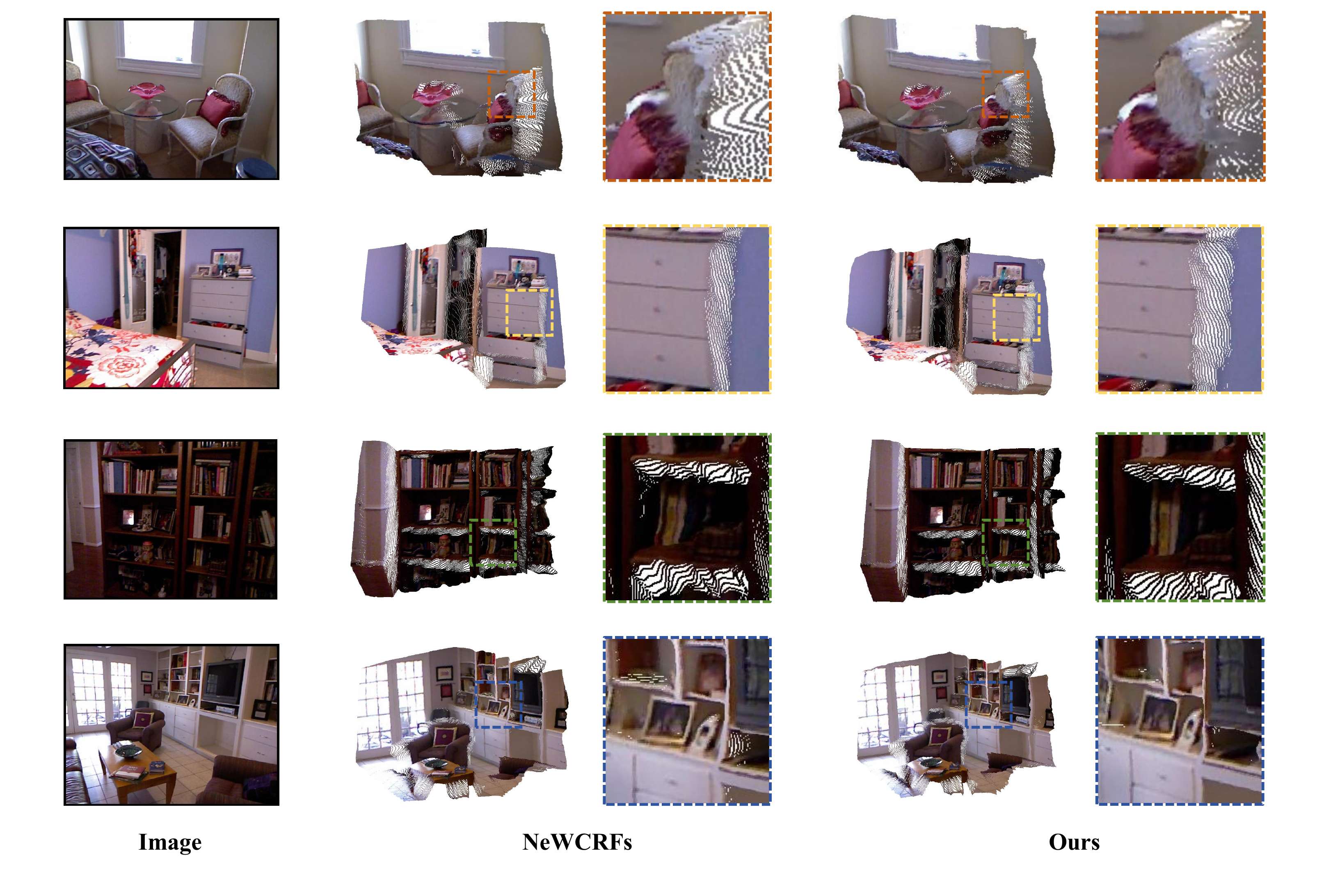} 
  \vspace{-3em}
  \caption{\textbf{The comparison of 3D reconstruction results on NYU-Depth V2 dataset.} The red bounding box region in the point cloud image is magnified and shown on the right side. We use camera intrinsic and the predicted depth maps to back-project the original image into point clouds.} 
  \label{fig:point_cloud} 
\end{figure*}

%% file: ablation2.tex
\begin{table}[t]
\centering
\caption{Ablation of number of transformer layers. AFM stands for adaptive feature modulator.}
\label{tab:4}
\begin{tabular}{c|c|ccc}
\toprule
Layer Number & AFM &  AbsRel & RMSE  & $\delta < 1.25$  \\ 
\midrule
3   & \ding{55} & 0.090  & 0.329 & 0.929 \\ 
6   & \ding{55} & 0.086  & 0.314 & 0.937 \\
9   & \ding{55} & 0.086  & 0.309 & 0.938  \\ 
\midrule
3(+2)   & \ding{51} & \textbf{0.084}  & \textbf{0.308} & \textbf{0.940} \\ 
\bottomrule
\end{tabular}
\end{table}

%% file: ablation1.tex
\begin{table}[t]
\centering
\caption{Ablation study of plane coefficient constraints. ``NC" stands for normal constraints, and ``DC" stands for distance constraints. }
\label{tab:5}
\begin{tabular}{c|cc|ccc}
\toprule
Row & NC & DC & AbsRel & RMSE  & $\delta < 1.25$  \\ 
\midrule
1   & \ding{55} & \ding{55} & 0.086  & 0.318 & 0.937 \\
2   & \ding{55} & \ding{51} & 0.086  & 0.319 & 0.936 \\
3   & \ding{51} & \ding{55} & 0.086  & 0.320 & 0.936 \\
4   & \ding{51} & \ding{51} & \textbf{0.084}  & \textbf{0.308}& \textbf{0.940}\\
\bottomrule
\end{tabular}
\end{table}

%% file: ablation_qnum.tex
\begin{table}[t]
\centering
\caption{Ablation study of numbers of plane queries. }
\label{tab:6}
\begin{tabular}{c|ccc|c}
\toprule
Numbers  & AbsRel & RMSE & log10  & $\delta < 1.25$  \\
\midrule
8    & 0.086 & 0.316  & 0.0371 & 0.938 \\
32   & \textbf{0.085} & \textbf{0.312}  & \textbf{0.0368} & \textbf{0.939} \\
100  & 0.086 & 0.313  & 0.0370 & 0.938 \\
128  & 0.087 & 0.314  & 0.0372 & 0.938 \\
\bottomrule
\end{tabular}

\end{table}